\documentclass[10pt,twocolumn,letterpaper]{article}

\usepackage{cvpr}              

\PassOptionsToPackage{table,xcdraw,dvipsnames}{xcolor}

\usepackage{adjustbox}
\usepackage{multirow} 
\usepackage{threeparttable}
\usepackage{amsmath}
\usepackage{amssymb}
\usepackage{mathtools}
\usepackage{bm}
\usepackage{tcolorbox}
\usepackage{enumitem} 
\usepackage{listings}

\usepackage{colortbl} 
\usepackage{arydshln}
\usepackage{pifont}

\usepackage{tikz}

\definecolor{bkcolor}{HTML}{4E79A7}
\definecolor{ourscolor}{HTML}{F28E2B}

\newcommand{\bx}{\mathbf{x}}

\definecolor{cvprblue}{rgb}{0.21,0.49,0.74}
\usepackage[pagebackref,breaklinks,colorlinks,allcolors=cvprblue]{hyperref}
\usepackage[capitalize]{cleveref}
\crefname{appendix}{Sec.}{Secs.}
\Crefname{appendix}{Sec.}{Secs.}
\usepackage{array} 
\usepackage{multirow}
\usepackage{amsmath}
\usepackage{amssymb}
\usepackage{mathtools}

\def\paperID{8810} 
\def\confName{CVPR}
\def\confYear{2025}

\title{Random Conditioning with Distillation for Data-Efficient \\Diffusion Model Compression}

\author{
Dohyun Kim$^{1}$\thanks{Equal contribution} \quad 
Sehwan Park$^{1}$\footnotemark[1] \quad 
Geonhee Han$^{1}$ \quad \\
Seung Wook Kim$^{2}$ \quad 
Paul Hongsuck Seo$^{1}$ \\
$^{1}$Dept. of CSE, Korea University \quad
$^{2}$NVIDIA \\
{\tt\small \{a12s12, shp216, rtrt505, phseo\}@korea.ac.kr}, \quad
{\tt\small seungwookk@nvidia.com}
}

\begin{document}
\maketitle
\begin{abstract}
Diffusion models generate high-quality images through progressive denoising but are computationally intensive due to large model sizes and repeated sampling.
Knowledge distillation—transferring knowledge from a complex teacher to a simpler student model—has been widely studied in recognition tasks, particularly for transferring concepts unseen during student training.
However, its application to diffusion models remains underexplored, especially in enabling student models to generate concepts not covered by the training images.
In this work, we propose Random Conditioning, a novel approach that pairs noised images with randomly selected text conditions to enable efficient, image-free knowledge distillation. 
By leveraging this technique, we show that the student can generate concepts unseen in the training images.
When applied to conditional diffusion model distillation, our method allows the student to explore the condition space without generating condition-specific images, resulting in notable improvements in both generation quality and efficiency.
This promotes resource-efficient deployment of generative diffusion models, broadening their accessibility for both research and real-world applications.
Code, models, and datasets are available at: \url{https://dohyun-as.github.io/Random-Conditioning}

\end{abstract}    
\section{Introduction}
\label{sec:intro}

\begin{figure}[t]
    \centering
    \includegraphics[width=1.0\linewidth]{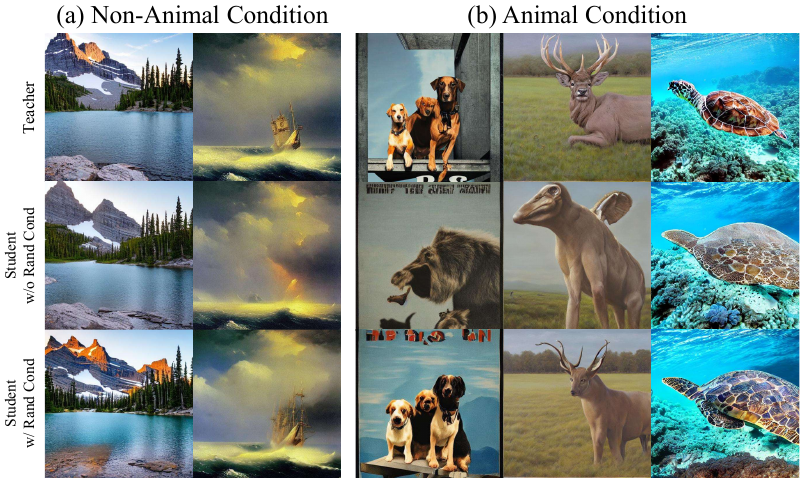}
    \caption{\textbf{Qualitative Comparison of Baseline and Our Method Trained Without Animal Image Data.} We train models on a dataset excluding animal-related images, both without and with random conditioning. Each row represents (from top to bottom) the teacher model, the model trained without random conditioning, and the model trained with random conditioning. In (a), samples are generated conditioned on captions unrelated to animals, and in (b), samples are generated conditioned on captions related to animals. The captions used to generate these samples are provided in~\cref{detail_prompts} of the Supp. Mat. for reference.}
    \label{fig:comp_seen_unseen}
    \vspace{-0.25cm}
\end{figure}

Diffusion models have emerged as powerful generative frameworks capable of producing high-quality outputs in various domains, such as image~\cite{ddpm, Imagen, dalle2, Sdm_v1.4, sdm_v1.5, LDM, SDXL}, video~\cite{alignyourlatents, animatediff, ImagenVideo, modelscope, videocrafter2, videodiffusionmodels}, and audio~\cite{audioldm2, auffusion, makeanaudio2}, by progressively denoising random noise through a sequence of learned steps.
Particularly, text-to-image diffusion models trained on large-scale datasets—such as Stable Diffusion~\cite {LDM, Sdm_v1.4, sdm_v1.5, SDXL}—excel at generating visually appealing images that accurately align with text prompts. Despite their impressive performance, these models come with significant computational demands, driven by a large number of sampling steps and extensive model parameters. Consequently, there has been growing interest in developing more efficient versions of these models. In this work, we focus on compressing conditional diffusion models to make them more efficient, especially in common real-life scenarios where access to large-scale data is limited, due to practical challenges such as hardware limitations, privacy concerns and licensing restrictions.

Knowledge distillation is a technique that transfers knowledge from one trained network, often a more complex model called the \textit{teacher}, to another, typically a simpler network known as the \textit{student}.
Through the use of soft targets~\cite{hinton2015distilling,soft_target1,soft_target2,soft_target3} or intermediate features~\cite{intermediate_distill1, intermediate_distill2, intermediate_distill3, intermediate_distill4} from the teacher model, which captures the relationships between concepts, 
distillation techniques are known to transfer not only seen but also unseen concepts to the student model.
For instance, \cite{hinton2015distilling} demonstrates that the student model learns to recognize the digit `3' on MNIST~\cite{mnist} although it was never provided an image of `3' during the distillation. 
Similarly, \cite{KD_unseen} provides a detailed analysis showing that a teacher’s knowledge across multiple domains can be transferred to a student model, even when distillation is performed using data from a single domain.
This capability to transfer knowledge of unseen concepts enhances the efficiency of training the student model, making it possible to achieve effective learning even with limited data.

\begin{figure}[t]
    \centering
    \includegraphics[width=1.0\linewidth]{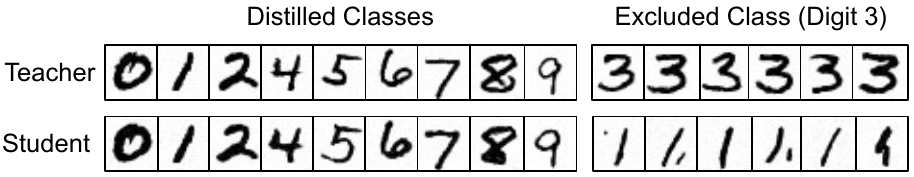}
    \caption{\textbf{Generated MNIST Images of Distilled and Excluded Digits by Teacher and Student.} 
    When the student is distilled using a dataset containing only a subset of digits, it fails to generate the excluded digit (`3’). Images from both the teacher and student models are generated with the same random seed for comparison.
    }
    \label{fig:unseen_failure}
\end{figure}

However, unlike in recognition models, this phenomenon is not observed in the context of
conditional diffusion models, as we demonstrate in \cref{sec:naive_approach} and \cref{fig:unseen_failure}.
The generative function in conditional diffusion models maps the semantic conditioning space to a much larger image space, making it harder for the student model to
generalize to unseen concepts.
The output noise is also specific to the current input, capturing minimal relationships across different output images. 
Additionally, each denoising step relies not only on the input condition but also on the intermediate noised images, which further complicates the mapping function.
As a result, it becomes challenging for the student model to infer unseen concepts effectively through distillation,
necessitating exploration of the entire conditioning space with a large set of condition-image pairs to fully distill the teacher model's generative capacity. 
However, acquiring such large-scale text-image pairs is often complicated by issues like copyright, privacy, and the storage constraints associated with handling image data.
Furthermore, even when images are generated using the teacher model from a text-only dataset, synthesizing them for all possible text prompts can be prohibitively expensive in terms of both computational resources and time.

To address these challenges, we propose a novel technique called random conditioning, where a noised image is paired with a randomly selected, potentially unrelated text condition during training.
This method allows the model to learn generalizable patterns without the need to generate images for every text prompt in the dataset, enabling efficient image-free distillation. 
By reducing the computational and storage demands associated with full image-text mappings, random conditioning preserves strong performance while significantly lowering resource requirements. 
Our preliminary experiments offer insights into the effectiveness of random conditioning, while extensive main experiments demonstrate that it enables student models to explore an extended condition space.
Consequently, as illustrated in \cref{fig:comp_seen_unseen}, the student learns to generate images containing unseen concepts (\eg, animals in \cref{fig:comp_seen_unseen}) even when images of these concepts are never provided during the distillation process.

Our main contributions are threefold:

\begin{itemize} 
    \item We provide a novel insight that conditional diffusion models fail to learn teacher knowledge for conditions that are not explicitly explored during the distillation process.
    \item We propose a novel technique, random conditioning, which allows the student model to explore conditions without requiring paired images.
    \item Leveraging this technique, we achieve efficient, image-free distillation of conditional diffusion models, producing compact models with competitive generative quality.
\end{itemize}

\section{Related Work}
\label{sec:relwork}

\noindent \textbf{Knowledge Distillation for Model Compression} \ \ 
Knowledge distillation is a common approach for model compression, where a smaller model learns to mimic the soft outputs~\cite{hinton2015distilling, soft_target1, soft_target2, soft_target3} or intermediate features~\cite{intermediate_distill1, intermediate_distill2, intermediate_distill3, intermediate_distill4} of a larger model, achieving significant compression with minimal performance loss. 
This technique has been effectively applied across various domains~\cite{panchapagesan2021efficient, KD_Object_Detection, wang2020minilm}, including large language models (LLMs)~\cite{distilbert, Sun2019PatientKD, jiao2020tinybert} and vision transformers (ViTs)~\cite{pmlr-v139-touvron21a, hao2022learning}, enabling the creation of models suitable for resource-constrained environments.
In these applications, student models successfully learn to generalize to inputs not explicitly exposed during distillation~\cite{hinton2015distilling,KD_unseen}.
However, in the context of conditional diffusion models, transferring knowledge for uncovered concepts through distillation remains underexplored.
Thus, we investigate this aspect within the scope of data-efficient model compression for diffusion models.

\noindent \textbf{Size-Reduced Diffusion Models} \ \ 
While diffusion-based generative models~\cite{LDM,sdm_v1.5,Sdm_v1.4,SDXL, chen2023pixartalpha,chen2024pixartdelta,chen2024pixartsigma} have shown strong performances, their large parameter counts and model sizes make them difficult to deploy in resource-constrained settings.
To address these challenges, various studies~\cite{chen2023speed, DKDM, fang2023structural} have focused on reducing model size through techniques such as quantization~\cite{so2023temporal, shang2023ptqdm}, architecture evolution~\cite{li2023snapfusion}, and knowledge distillation~\cite{kim2023bksdm, Lee@koala}. Notably, BK-SDM~\cite{kim2023bksdm} compresses stable diffusion~\cite{Sdm_v1.4,sdm_v1.5} into smaller versions by applying block pruning and feature distillation while KOALA~\cite{Lee@koala} compresses SDXL~\cite{SDXL,sauer2023sdxl-turbo,lin2024lightning} by employing layer-wise removal and self-attention-based knowledge distillation.
We build on previous studies by analyzing the effectiveness of knowledge distillation in conditional diffusion models and propose a general approach for more efficient distillation of diffusion models.

\noindent \textbf{Diffusion Acceleration} \ \ 
Recent studies on accelerating diffusion models have focused on reducing the number of sampling steps, rooted in the iterative refinement process of diffusion models.
A line of studies aims at accelerating denoising process in diffusion models without training~\cite{karras2022elucidating,lu2022dpm, zheng2023dpmsolverv}, resulting in dramatically reduced sampling steps from a thousand to 10–25.
However, further reductions often cause a steep decline in performance. 
Distillation-based accelerating methods~\cite{salimans2022progressive, meng2023distillation, song2023consistency,kim2023consistency, luo2023latent_Consistency, DMD, DMD2, liu2023instaflow,liu2022flow, BOOT} approach this challenge through knowledge distillation, enabling student models to consolidate multi-step outputs into single-step predictions. For example, Consistency Distillation~\cite{song2023consistency,kim2023consistency, luo2023latent_Consistency} trains models to produce self-consistent outputs across timesteps, facilitating accurate single-step predictions. These works do not focus on compressing model size; instead, they aim to create few- or one-step models based on a base model. Our research, on the other hand, aims to develop a compressed base model, which could serve as a complementary foundation for step-acceleration methods, enhancing their effectiveness.

\begin{figure*}[t]
    \centering
    \includegraphics[width=1\linewidth]{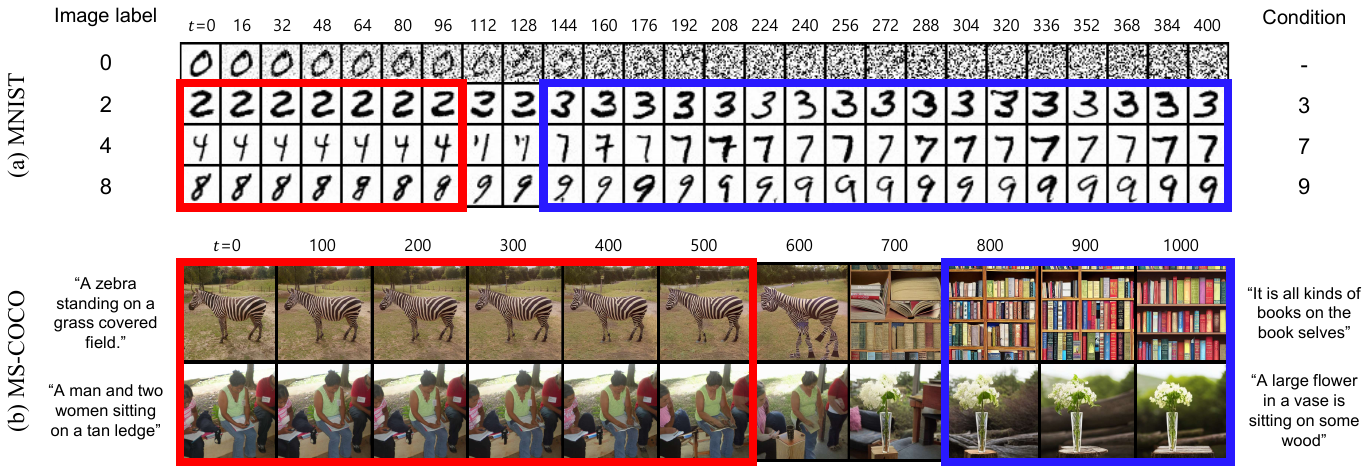} 
    \caption{\textbf{Effects of Altered Conditioning on Generated Results from an Input Image across Timesteps.} Generated results conditioned on the rightmost column using the input image from the leftmost column at each timestep for both MNIST \cite{mnist} and MSCOCO \cite{mscoco}. First, $\bx_t$ is derived from the initial image $\bx_0$, associated with the image label, at the timestep $t$ shown above each image using the forward process and then, $\bx_0$ is regenerated through the reverse process, conditioned on the displayed rightmost column.}
    \label{fig:outputs_per_step}
\end{figure*}

\section{Method}
\label{sec:method}

In this section, we present our novel approach for distilling conditional diffusion models into smaller student models. \cref{sec:task_definition} outlines the problem we aim to address and the associated challenges encountered in this process. \cref{sec:naive_approach} describes a na\"ive  baseline approach to tackle this problem, while \cref{sec:random_conditioning} introduces our proposed method called random conditioning, including its motivation and key observations.

\subsection{Distilling Diffusion Models for Compression}
\label{sec:task_definition}
Our task is to compress a conditional diffusion model, and in this work we showcase this with Stable Diffusion model for text-to-image generation~\cite{Sdm_v1.4, sdm_v1.5, LDM, SDXL}, as it is one of the most widely used conditional diffusion models. In other words, we distill the knowledge within a teacher diffusion model $\mathcal{T}$ trained at scale to an arbitrary student model $\mathcal{S}$ that can have a different architecture with a significantly smaller number of parameters. 
Notably, this task differs from diffusion acceleration via knowledge distillation~\cite{salimans2022progressive, meng2023distillation, song2023consistency,kim2023consistency, luo2023latent_Consistency, DMD, DMD2, liu2023instaflow,liu2022flow, BOOT}, where the primary aim is to distill a model to decrease the number of diffusion steps required for inference.
We approach this task in an image-free setting, where only text prompts are available, without access to any images.
This configuration is especially useful, as collecting large-scale image-text pairs is challenging. 
The process is costly, requires intensive labor for accurate annotation, and is further complicated by privacy concerns and licensing restrictions, which limit access to diverse, high-quality datasets.
In certain domains, these issues are even more pronounced, where data scarcity or heightened privacy concerns make it especially difficult to obtain well-annotated image-text pairs.

Applying knowledge distillation to diffusion models without images introduces additional challenges due to the iterative nature of the denoising process.
In diffusion models, the forward and reverse processes are defined on some time interval $[0, T]$, and the teacher model predicts the noise $\boldsymbol{\epsilon}_\mathcal{T}(\bx_t,t,c)$ to be removed from $\bx_t$ at each timestep $t \in [0,T]$ given a text condition $c$.
Therefore, knowledge transfer from the teacher model to the student model must occur at each timestep $t$.
However, without access to images, generating the intermediate noisy input $\bx_t$, which is typically created by adding noise to the original image $\bx_0$~\cite{thermodynamics, ddpm}, becomes challenging. 
This limitation prevents us from performing knowledge distillation for $t\ne T$ where $T$ is the total number of denoising steps, as we lack the necessary input image at intermediate timesteps.

\subsection{Na\"ive Baseline Approach}
\label{sec:naive_approach}

A na\"ive approach for image-free distillation would involve generating images for all available text prompts to construct a paired dataset $\mathcal{D}=\{(\bx^n, c^n)\}^N_{n=1}$ where $\bx^n$ is the generated image that serves as original image $\bx_0$ for the text condition $c^n$ allowing us to construct noisy input image $\bx_t$ for any timestep $t$ and condition $c^n$.
Since diffusion models are time-intensive for image generation, we need to generate and cache these images in advance to build the dataset.
The teacher model can then be distilled into a student model with the following loss function:
\begin{align}
\mathcal{L}_{\text{out}} &= \mathbb{E}_{(\bx_t,c)\in \mathcal{D}, t}\left[\|\boldsymbol{\epsilon}_\mathcal{T}(\bx_t, c, t) - \boldsymbol{\epsilon}_\mathcal{S}(\bx_t, c, t)\|_2^2\right],
\label{eq:img_loss}
\end{align}
where $\boldsymbol{\epsilon}_\mathcal{T}$ and $\boldsymbol{\epsilon}_\mathcal{S}$ are the predicted noises by the teacher and student models, respectively.
Here, $(\bx_t,c)$ is a pair sampled from the dataset $\mathcal{D}$ with noise injected into the image based on $t$, which is uniformly distributed between 0 and $T$.
In addition, we may incorporate a feature-level knowledge distillation loss function, which is given by
\begin{align}
\mathcal{L}_{\text{feat}} = \mathbb{E}_{(\bx_t,c)\in \mathcal{D}, t}\left[\sum_l \|\mathbf{f}_\mathcal{T}^l(\bx_t, c, t) - \mathbf{f}_\mathcal{S}^l(\bx_t, c, t)\|_2^2\right], 
\label{eq:feat_loss}
\end{align}
where $\mathbf{f}_\mathcal{T}^l$ is the feature maps from layer $l$ of the teacher model and $\mathbf{f}_\mathcal{S}^l$ denotes the feature maps from the corresponding layer of the student models. 
Note that $\mathcal{T}$ and $\mathcal{S}$ do not need to have the same architecture;
we can incorporate additional temporary modules for distillation to project arbitrary intermediate features of $\mathcal{S}$ to $\mathbf{f}_\mathcal{S}^l$ with the same dimensionality as the corresponding features $\mathbf{f}_\mathcal{T}^l$.
These additional projection modules are then discarded after the distillation process.
This feature-level loss, combined with the noise prediction loss, encourages the student model to replicate both the external outputs and the internal processing of the teacher model.
It allows the student to learn and replicate the teacher model’s denoising behavior for the text conditions $c$ encountered during the distillation process.

While this naïve approach enables effective knowledge transfer from the teacher to the student model, it presents several limitations. The method requires generating images $\bx_0$ for a diverse set of text prompts to sufficiently cover the text condition space. 
Without covering the entire condition space, the student model may fail to generate images for those conditions that have never been observed during distillation.
Our preliminary experiment on MNIST~\cite{mnist} in~\cref{fig:unseen_failure} illustrates the importance of covering the condition space.
Although the teacher model can generate the digit `3', the student model fails to produce this digit when it has not been exposed to this condition during distillation.
Since the text condition space is exceedingly large—unlike the 10-digit space in MNIST—synthesizing $\bx_0$ for all possible prompts becomes prohibitively costly in terms of computation, time, and storage. 
This challenge is particularly significant with diffusion models, which rely on multiple timesteps during inference, further compounding the computational demands for each generated image.

\begin{figure}[t]
    \centering
    \includegraphics[width=1\linewidth]{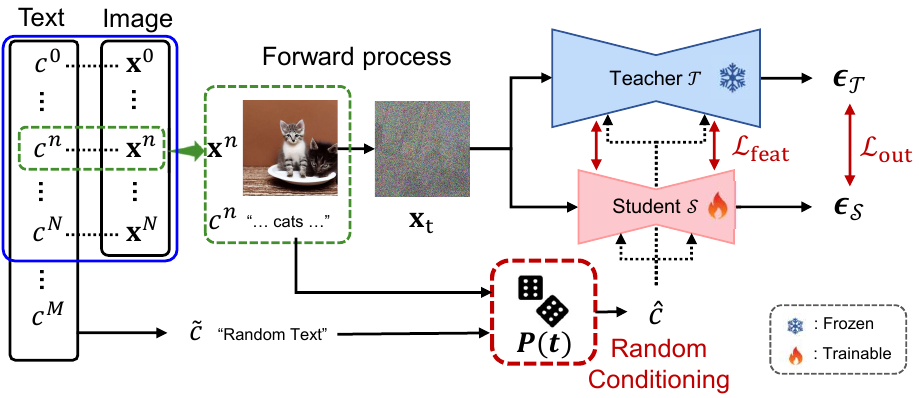}
    \caption{\textbf{Overview of the Random Conditioning Approach.} When distilling knowledge from the teacher model to a smaller student model, instead of pairing each training image dataset sample $\bx_{t}^{n}$ with its original condition $c^n$, we replace it with a random condition $\Tilde{c}$ from the text dataset based on a predefined probability $p(t)$ at each timestep $t$. This approach enables the student model to learn the teacher's behavior even for conditions without explicit image pairs.}
    \label{fig:random_conditioning}
\end{figure}

\subsection{Random Conditioning}
\label{sec:random_conditioning}
To address the above challenges, we propose random conditioning illustrated in~\cref{fig:random_conditioning} that allows us to cache images generated from only a subset of text prompts (blue box).
Formally, given an extensive set of $M$ text prompts $\mathcal{C}$, we construct a dataset of $N$ image-text pairs $\mathcal{D}={(\bx^n, c^n)}$ where $N\ll M$.
As discussed above, training a student model on this paired dataset $\mathcal{D}$ would limit the knowledge transfer in distillation as there are many uncovered parts in the text condition space that could be covered by those texts in $\mathcal{C}$.
Note that this limitation arises from the absence of noisy input images $\bx_t$, which are typically constructed from the original image $\bx_0$, with the generated images in $\mathcal{D}$ serving as these originals.
In our approach, we leverage not only $\mathcal{D}$, which contains a limited number of generated images, but also $\mathcal{C}$, allowing the student model to explore all text conditions in $\mathcal{C}$. 
This approach enhances the distilled knowledge, enabling the model to generalize across the full condition space.

Precisely, we first sample a paired data $\bx^n$ and $c^n$ from $\mathcal{D}$ and construct $\bx_t$ from $\bx^n$.
Then, before performing distillation, we apply a predefined random conditioning probability $p(t)$ to sample a random text from $\mathcal{C}$.
Specifically, the text condition $\hat{c}$ is determined by
\begin{equation}
    \hat{c} = 
    \begin{cases} 
          c^n & \text{with probability } 1 - p(t), \\
          \Tilde{c} \in  \mathcal{C} & \text{with probability } p(t),
    \end{cases}
\end{equation}
where $\Tilde{c}$ is randomly sampled from $\mathcal{C}$.
Finally, $\hat{c}$ is paired with $\bx_t$ to compute both distillation losses defined in~\cref{eq:img_loss} and \eqref{eq:feat_loss}.

\begin{figure}[t]
    \centering
    \includegraphics[width=1\linewidth]{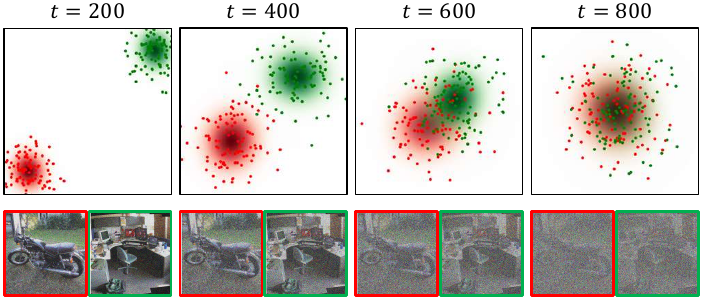}
    \caption{\textbf{Distributions of $p(\bx_t|c^n)$ and $p(\bx_t|\Tilde{c})$.} Visualization of the distributions of toy 2D data samples at timesteps 200, 400, 600, and 800, along with corresponding $\bx_t$ images at each timestep. As the timestep increases, the distributions progressively overlap with each other.}
    \label{fig:dist_overlap}
\end{figure}

\noindent \textbf{Observations and Motivation} \ \ 
While the proposed random conditioning technique may initially appear counterintuitive, it is grounded in our empirical observation that diffusion models incorporate conditioning information in a manner that varies with the timestep $t$. \cref{fig:outputs_per_step} shows the generated outputs during the denoising process, starting from $\bx_t$ at various timesteps $t$ on the MNIST~\cite{mnist} and MS-COCO~\cite{mscoco} datasets.
In each row, $\bx_t$ is derived from the same initial image $\bx_0$ corresponding to the leftmost column, and the generated outputs share the same conditioning, displayed in the rightmost column. 
Notably, this condition differs from the label associated with the original image $\bx_0$.
The generated images primarily align with either the original image label or the conditioning value, with only a narrow range of $t$ producing outputs with noticeable artifacts. 
Specifically, when $t$ is small, the generated images tend to reflect the original image label (red boxes) due to the low noise magnitude characteristic of later steps in the denoising process. 
Conversely, when $t$ is large, the generated images predominantly follow the conditioning value (blue boxes), as the input $\bx_t$ becomes nearly indistinguishable from pure noise.
These results also indicate that the condition $c$ does not need to be strongly correlated with the noised input $\bx_t$ supporting the proposed random conditioning technique.
This is due to: (1) the model’s tendency to rely almost entirely on the input condition $c$ at large $t$ where the original semantics of $\bx_0$ are nearly lost, and (2) the model’s primary focus on denoising the input $\bx_t$ while disregarding the condition $c$ when $t$ is small.
Furthermore, \cref{fig:dist_overlap} demonstrates that, as the noise level or timestep $t$ increases during the forward process, the distributions $p(\bx_t|c^n)$ and $p(\bx_t|\Tilde{c})$ become closer to each other, eventually merging into the same Gaussian distribution as $t$ approaches $T$. This observation implies that the input image and condition do not need to be directly aligned at every timestep. It supports both the effectiveness and validity of our random conditioning method, highlighting its flexibility in associating conditions with diverse inputs.
Based on these observations and the motivation, we empirically explored $p(t)$. When $p(t)$ was set as a constant value, such as $p(t)=1$, the results were suboptimal. In particular, reducing $p(t)$ for intermediate time steps, where the pairing between the image and condition becomes relatively more important, led to improved performance. Among these, we used an exponential function for $p(t)$ in our experiments. Further experiments regarding $p(t)$ are provided in~\cref{sup p test} of the Supp. Mat.

\noindent \textbf{Extended Exploration of Condition Space} \ \ 
As explored in~\cref{fig:unseen_failure}, the student model effectively learns to generate images for conditions explicitly covered by the paired dataset $\mathcal{D}$ during distillation, 
but generating images for every text prompt in $\mathcal{C}$ poses a significant bottleneck.
Random conditioning alleviates this by allowing the use of conditions not included in $\mathcal{D}$ to be applied without requiring paired images.
Consequently, the student can explore text prompts beyond those paired with images, even when the number of conditions far exceeds the available images.
This setup helps the student replicate the teacher's behavior under novel conditions, thereby broadening its generative capabilities.

\section{Experiments}
\label{sec:Experiments}

\subsection{Datasets}

\noindent \textbf{LAION} \ \ 
We use LAION~\cite{laion400M, laion_aesthetics} consisting of 400M image-text pairs.
Following~\cite{kim2023bksdm}, we use 212K samples from the LAION-Aesthetics V2 (L-Aes) 6.5+~\cite{laion_aesthetics}, which is a subset of LAION.
To simulate image-free training, we extract text prompts only from those 212K samples and generate their images.
For random conditioning, we use 20M extra text prompts randomly sampled from 400M pairs of LAION.
It is worth noting that original images from LAION are still used in baseline methods, as these methods are developed for setups requiring image access.

\noindent \textbf{MS-COCO} \ \ 
The MS-COCO~\cite{mscoco} dataset is a large-scale text-image paired dataset with diverse and detailed annotations, including 80 object classes. 
Following previous practices~\cite{dalle1, Imagen, LDM}, we use 30K image-text pairs sampled from the MS-COCO validation split, which consists of 41K images, each with five human-annotated captions. 
For each image, we use a single caption preselected from~\cite{kim2023bksdm} for evaluation.

\subsection{Experimental Settings}
\label{sec:exp_settings}
\noindent \textbf{Models} \ \ 
For our experiments, we use the Stable Diffusion (SD) v1.4 model~\cite{Sdm_v1.4} as the teacher model. 
BK-SDM~\cite{kim2023bksdm} serves as our baseline method, representing the current SOTA in diffusion model compression using knowledge distillation. 
Unlike our approach, which operates in an image-free setup, BK-SDM utilizes both the original images and text.
BK-SDM proposes three compressed architectures—Base, Small, and Tiny—by selectively removing blocks from the teacher network to achieve compression. 
For a fair comparison, we evaluate our method using the same compressed architectures.
Additionally, we evaluate four further compressed architectures that reduce the number of channels while preserving all layers, offering an alternative compression approach. 
Three of these architectures are sized to match the Base, Small, and Tiny configurations from BK-SDM, while the fourth is even smaller than the Tiny architecture (C-Micro), pushing compression further.

\noindent \textbf{Evaluation Metrics} \ \ 
We evaluate the models using standard metrics commonly applied in text-to-image generation: 
Fréchet Inception Distance (FID)~\cite{FID}, Inception Score (IS)~\cite{IS}, and the CLIP score~\cite{clipscore, clipscore2}. 
FID and IS focus on measuring the visual fidelity and diversity of generated images. 
The CLIP score evaluates the alignment between the generated image and the text prompt. 
We use the Inception-v3 model for computing FID and IS, while the ViT-g/14 model is used for calculating the CLIP score.

\noindent \textbf{Implementation Details} \ \ 
We adopt all hyperparameters from~\cite{kim2023bksdm} except for null condition proportion~\cite{cfg}, which is set to 10\%.
We utilize four 40GB NVIDIA A100 GPUs with a batch size of 256 for distillation. 
We train the models using the AdamW~\cite{AdamW} optimizer with a learning rate of 5e-5. 
We use the two losses of~\cref{eq:img_loss} and~\cref{eq:feat_loss}, with equal weights of 1.
For~\cref{eq:feat_loss}, feature distance is reduced after each block in the U-Net~\cite{unet, beatGans}.

\subsection{Results}

\label{sec:Experimental_Results}

\begin{table}[t]
  \centering
\scalebox{0.85}{
  \begin{tabular}{ccccccc}
    \hline
    \hline
    \# & Rand Cond & T Init & Real image & FID↓ & IS↑ & CLIP↑ \\
    \hline
    1 & \ding{55} & \ding{55} & \ding{55}  & 18.13 & 31.84 & 0.2728\\
    2 & \ding{55} & \ding{51} & \ding{55}  & 18.15 & 33.81 & 0.2864 \\
    3 & \ding{55} & \ding{51} & \ding{51}  & 15.76 & 33.79 & 0.2878 \\
    \hline
    4 & \ding{51} & \ding{55} & \ding{55}  & 15.46 & 34.48 & 0.2834 \\
    5 & \ding{51} & \ding{51} & \ding{55}  & 15.76 & 36.03 & 0.2895 \\
    6 & \ding{51} & \ding{51} & \ding{51}  & \textbf{15.00} & \textbf{36.14} & \textbf{0.2933} \\
    \hline
    \hline
  \end{tabular}
  }
  \caption{\textbf{Impact of Random Conditioning.} We compare models trained with and without random conditioning across various settings, varying teacher initialization and the availability of real images on MS COCO-30k. All models are based on the B-Base architecture. “Rand Cond” denotes whether random conditioning is applied, “T Init” indicates whether the model is initialized from the teacher model, and “Real image” specifies the use of real images during training. Notably, Row~3 is the same as BK-SDM~\cite{kim2023bksdm}.}
  \label{tab:RC_effect}
\end{table}

\begin{table*}[t]
    \centering
    \scalebox{0.85}{
    \begin{tabular}{
        c  
        c  
        c  
        c  
        c   
        c  
        c   
        c   
        c  
        c   
        c   
        c    
    }
        \hline
        \hline
        & & & 
        \multicolumn{3}{c}{Seen (Non-animal)} & 
        \multicolumn{3}{c}{Unseen (Animal)} &  
        \multicolumn{3}{c}{Seen+Unseen} \\
        \cline{4-12}
        \# & Rand Cond & Additional Texts & 
        FID\(\downarrow\) & IS\(\uparrow\) & CLIP\(\uparrow\) & 
        FID\(\downarrow\) & IS\(\uparrow\) & CLIP\(\uparrow\) & 
        FID\(\downarrow\) & IS\(\uparrow\) & CLIP\(\uparrow\) \\
        \hline
        \rowcolor{gray!25} 
        \multicolumn{3}{c}{(Teacher)} & 
        13.29 & 32.47 & 0.2954 & 
        22.53 & 18.63 & 0.3035 & 
        12.67 & 36.71 & 0.2971\\
        \hline
        1 & 
        \ding{55} & None  & 
         15.24 & 28.11 & 0.2801 & 
         37.86 & \textbf{17.73}  & 0.2478 & 
         15.66 & 29.62 & 0.2734  \\
        2 & 
        \ding{51} & 24K animal-related texts & 
         \textbf{14.42} & 27.86 & 0.2788  & 
         \textbf{23.26} & 17.18 & 0.2833 & 
         \textbf{13.50} & 31.30 & 0.2797 \\
        3 & 
        \ding{51} & 24K+20M& 
         15.37 & \textbf{30.27} & \textbf{0.2879} & 
         24.71 & 17.39 & \textbf{0.2913} &
         14.47 & \textbf{34.06} & \textbf{0.2886}\\
        \hline
        \hline
    \end{tabular}
    }
    
    \vspace{-0.2cm}
    \caption{
    \textbf{Knowledge Transfer of Unseen Concepts through Random Conditioning.}
    The row with a gray background shows the performance of the teacher model~\cite{Sdm_v1.4} for reference.
    “Rand Cond” indicates the use of random conditioning, and “Additional Texts” specifies the amount and source of extra text data used for random conditioning.
    All models are trained on images generated from approximately 188K non-animal prompts, obtained by excluding 24K animal-related samples from the original 212K LAION dataset.
    In addition,
    Row~2 leverages those 24K animal-related texts excluded from the set used to generate training images, while Row~3 further utilizes 20M LAION texts.
    Evaluation is conducted across three setups: “Seen,” which tests generation of non-animal concepts; “Unseen,” for animal-related concepts; and “Seen+Unseen,” which includes both.
    All student models use the B-Base architecture.
    }
    \label{tab:seen_unseen}
\end{table*}

\noindent \textbf{Effects of Random Conditioning}
\cref{tab:RC_effect} demonstrates the effectiveness of random conditioning. 
The top three rows show scores without random conditioning, while the bottom three rows display their corresponding scores with random conditioning applied.
In particular, comparing Rows 1 and 4, random conditioning shows a substantial performance boost with 14.72\% decrease in FID and 8.29\% increase in IS, indicating its significant impact.
Previous studies~\cite{kim2023bksdm} have demonstrated that initializing the student with teacher weights can enhance performances. 
Here, by comparing Rows 1 and 2, as well as Rows 4 and 5, we observe a similar performance increase through initialization. 
Notably, even when teacher initialization is applied, random conditioning still adds meaningful performance gains. 
Furthermore, models with random conditioning and random initialization achieve comparable or even superior performance to those with teacher initialization but without random conditioning, highlighting the powerful impact of random conditioning on model performance.

In Rows 5 and 6, the scores are nearly identical, underscoring that our method maintains strong performance without real image usage. Random conditioning contributes more significantly to achieving high scores than the impact of real image usage, establishing it as the key factor in score enhancement. By efficiently distilling knowledge from the teacher model, our approach achieves comparable results without needing real images, providing a practical and robust solution for knowledge distillation in conditional diffuion model even without access to actual image data.

\noindent \textbf{Knowledge Transfer of Unseen Concepts} \ \  
To demonstrate the effect of random conditioning in transferring knowledge of unseen concepts—specifically, conditions excluded from the paired training dataset $\mathcal{D}$—we train student models on a dataset that omits all images containing animals as the unseen concepts.
To exclude animal images from training, we apply a filtering process to the original 212K LAION~\cite{laion_aesthetics} dataset using GPT~\cite{openai2024gpt4technicalreport}, BLIP~\cite{blip}, and keyword elimination.
This process yields 188K non-animal text prompts from the original 212K samples.
The models are trained with generated images from these 188K prompts and for those with random conditioning, we utilize additional text prompts without generating images.
For evaluation, we test models on two subsets—33K non-animal prompts and 8K animal-related prompts—as well as on the MS-COCO 30K. Detailed process of filtering animal-related data from both the training and evaluation sets is provided respectively in  \cref{sec: detail filtering} of the Supp. Mat.

\cref{tab:seen_unseen} illustrates the results in this configuration.
Without random conditioning (Row 1), this model fails to learn the unseen concepts of animals showing poor performances whereas it maintains comparable performances in generating images with seen concepts.
When random conditioning is applied, the model (Row 2) achieves significant improvements especially in FID and the CLIP score by facilitating the filtered 24K texts but without their images. 
Finally, extending the text dataset with prompts from LAION (Row 3) leads to further improvements in both IS and CLIP scores for both seen and unseen cases. 

Beyond unseen categories, our model also surpasses those Base models without random conditioning in all metrics for seen concepts, achieving scores that closely approach those of the teacher model.
This suggests that random conditioning not only enhances knowledge of unseen concepts
but also boosts overall generation quality. Among our models, those utilizing more text data generally exhibit better performance overall. 
Qualitative results in~\cref{fig:comp_seen_unseen} further illustrate that the quality of generated images for unseen concepts is distinctly better when random conditioning is applied, compared to when it is not.
Detailed analysis of the impact of extra text dataset sizes and more qualitative examples are provided in~\cref{text size,sec_sup:qualitative results} of the Supp. Mat.

\begin{table}[t]
    \centering
    \scalebox{0.85}{
    \begin{tabular}{cccccc}
        \hline
        \hline
        \# & Rand Cond & Data Source & FID↓ & IS↑ & CLIP↑ \\
        \hline
        \rowcolor{gray!25} 
        \multicolumn{3}{c}{(Teacher)} & 13.05 & 36.76 & 0.2958\\
        \hline
        1 & \ding{55} &LAION & 18.15 & 33.81 & 0.2864\\
        2 & \ding{51} &LAION & 15.76 & 36.03 & 0.2896\\
        \hline
        3 & \ding{51} &GPT & \textbf{14.98} & \textbf{36.70} & \textbf{0.2952}\\
        \hline
        \hline
    \end{tabular}
    }
    \vspace{-0.1cm}
    \caption{
    \textbf{Model Comparisons with Varying Data Constraints.}
    The row with a gray background shows the performance of the teacher model~\cite{Sdm_v1.4} for reference.
    “Rand Cond” indicates whether random conditioning is used, and “Data Source” specifies the text data used for both paired image generation and additional conditioning.
    Row3 represents the fully data-free configuration, where even text data are unavailable and are generated automatically using an LLM (\eg, GPT).
    For fair comparison, all models are trained with 212K generated images.
    Row2 uses additional 20M LAION captions, while Row~3 uses 2.2M GPT-generated prompts.
    All student models are based on the B-Base architecture.}
    \vspace{-0.2cm}

    \label{tab:gpt_training}
\end{table}
\noindent \textbf{Data-Free Distillation} \ \ 
In \cref{tab:gpt_training}, we evaluate the effectiveness of random conditioning in a fully data-free setup, where even text data are unavailable for distillation.
In this setting (Row~3), text prompts are automatically generated by an LLM, as described in~\cref{Appendix:Data-Efficient_Distillation} of the Supp. Mat., and a 212K subset is used to synthesize images, forming a paired dataset.
Remarkably, even without real text data, the model in Row~3 not only outperforms the baseline without random conditioning (Row~1) but also achieves performance comparable to the model trained with real text data and random conditioning (Row~2). This demonstrates the scalability and adaptability of our method in resource-constrained settings.
Furthermore, LLM-generated captions in this setup can be tailored to the target domain, offering the potential to steer the student model toward specific generation styles or tasks.

\noindent \textbf{Comparisons to Other Text-to-Image Models} \ \ 
We build our models by applying two compression strategies: 
block compression, which removes UNet blocks, and channel compression, which reduces channel widths.
Block-compressed models (B-Base, B-Small, B-Tiny) follow~\cite{kim2023bksdm}, use pretrained teacher weights, and achieve significant parameter reduction with minimal performance drops.
Channel compression allows greater flexibility for higher compression rates.
We design C-Base, C-Small, C-Tiny with parameter counts comparable to block-compressed models, and introduce C-Micro, which has 30\% fewer parameters than B-Tiny.
Due to channel size mismatches, channel-compressed models cannot reuse teacher weights. Details on multiply-accumulate operations (MACs), UNet parameter counts, and additional comparisons between these models are provided in~\cref{sup compression} of the Supp. Mat. 

\begin{table}[t]
\centering
\scalebox{0.843}{
\setlength{\tabcolsep}{3.4pt}
\begin{tabular}{lccccc}
\hline
\hline
Models & \#Params\textsuperscript & \#Images &FID↓ & IS↑ & CLIP↑ \\ 
\hline
\rowcolor{gray!25}
SDM-v1.4~\cite{Sdm_v1.4, LDM}\textsuperscript{$\dagger$} & 1.04B & $>$2000M & 13.05 & 36.76 & 0.2958 \\ 
Small SD~\cite{ofasys}\textsuperscript{$\dagger$} & 0.76B & 229M & 12.76 & 32.33 & 0.2851 \\
BK-SDM-Base \textsuperscript{$\dagger$} & 0.76B & 0.22M& 15.76 & 33.79 & 0.2878 \\
BK-SDM-Small \textsuperscript{$\dagger$} & 0.66B & 0.22M& 16.98 & 31.68 & 0.2677 \\
BK-SDM-Tiny \textsuperscript{$\dagger$} & 0.50B & 0.22M&17.12 & 30.09 & 0.2653 \\ 
\hline
\rowcolor[HTML]{ECF4FF} 
B-Base [Ours] & 0.76B & 0 &14.47 & 36.50 & 0.2932\\
\rowcolor[HTML]{ECF4FF} 
B-Small [Ours]& 0.66B & 0& 16.22& 35.99 & 0.2804\\
\rowcolor[HTML]{ECF4FF} 
B-Tiny [Ours] & 0.50B & 0 & 16.71 & 35.46 & 0.2782\\
\rowcolor[HTML]{ECF4FF} 
C-Base [Ours] & 0.73B & 0 & 14.45 & 34.92 & 0.2904\\
\rowcolor[HTML]{ECF4FF} 
C-Small [Ours] & 0.61B & 0 &14.43  & 34.58 & 0.2888 \\  
\rowcolor[HTML]{ECF4FF} 
C-Tiny [Ours] & 0.49B& 0 & 13.90  & 33.18 & 0.2860 \\
\rowcolor[HTML]{ECF4FF} 
C-Micro  [Ours]& 0.40B & 0 & 13.42 & 32.64 & 0.2813\\

\hline
GLIDE~\cite{glide}\textsuperscript{$\dagger$} & 3.5B & 250M& 12.24 & - & - \\
LDM-KL-8-G~\cite{LDM}\textsuperscript{$\dagger$}& 1.45B & 400M& 12.63 & 30.29 & - \\
DALL·E-2~\cite{dalle2}\textsuperscript{$\dagger$} & 5.2B & 250M& 10.39 & - & - \\
SnapFusion~\cite{li2023snapfusion}\textsuperscript{$\dagger$} & 0.99B & $>$100M & $\sim$13.6 & - & $\sim$0.295 \\
Würstchen-v2~\cite{wuerstchen}\textsuperscript{$\dagger$} & 3.1B & 1700M & 22.40 & 32.87 & 0.2676  \\
Pixart-alpha~\cite{chen2023pixartalpha} & 5.4B & 25M & 23.43 & 34.54 & 0.3072 \\
SDXL-Base-1.0~\cite{SDXL} & 3.5B & - & 12.15 & 35.12 & 0.3199 \\
SD 3.5 Medium~\cite{SD3.5-Medium} & 7.9B & - & 16.23 & 39.81 & 0.3246 \\

\hline
\hline
\end{tabular}
}

\caption{\textbf{Comparison with Other Models on MS-COCO 30K.} Despite having significantly fewer parameters than other large models, our model achieves comparable performance with minimal quality degradation. ``\#Params'' refers to the total number of parameters.
``\#Images'' refers to the quantity of real images used in training. 
\textsuperscript{$\dagger$}Results reported from \cite{kim2023bksdm}.
}

\label{table:main_comp}
\end{table}

\begin{figure}[t]
    \centering
    \includegraphics[width=1.0\linewidth]{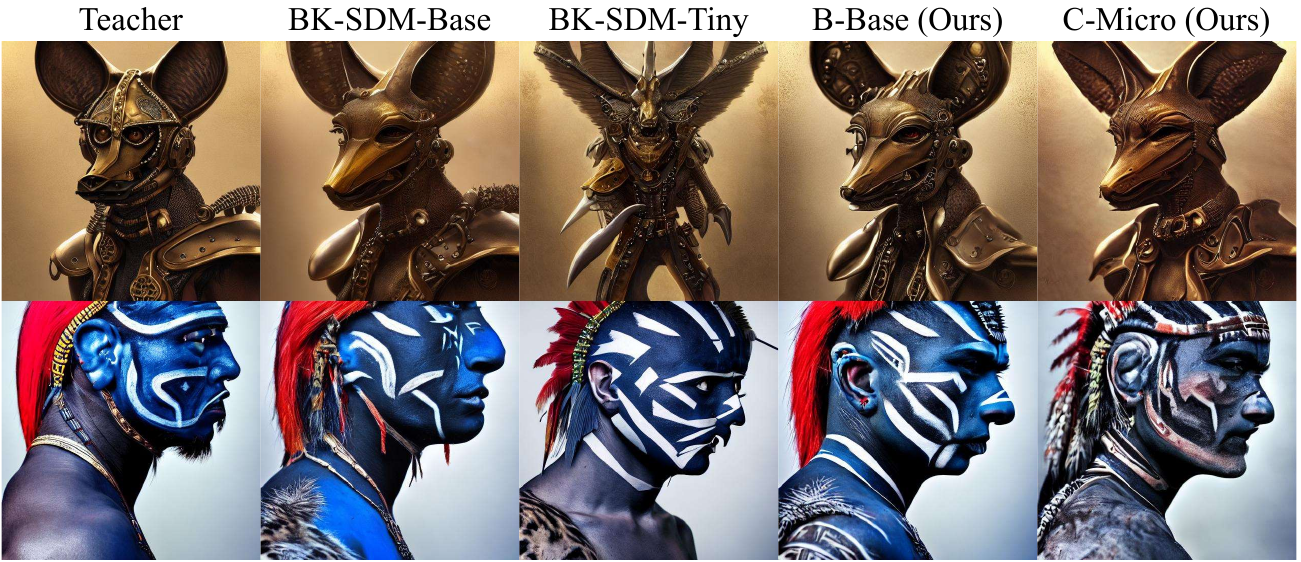}
    \caption{\textbf{Qualitative Comparison between Our Models and Baseline Models.} From left to right: samples generated from the teacher model, BK-SDM Base, BK-SDM Tiny, B-Base (ours), and C-Micro (ours) using the same prompts and seeds. The captions used are provided in~\cref{detail_prompts} of the Supp. Mat.}
    \label{fig:comp_others}
    \vspace{-0.2cm}
\end{figure}

\cref{table:main_comp} compares our compressed models with other diffusion models, presenting total parameter counts, number of real images used for training, and performances.
Our B-Base, B-Small, and B-Tiny models share the same architecture as their corresponding BK-SDM models and are distilled from the same teacher model. 
However, our enhanced distillation approach yields superior performance.
Note that our models with higher compression rates outperform BK-SDM’s larger models, despite BK-SDM being trained with real images and teacher initialization. 
For example, our smallest model, C-Micro, surpasses BK-SDM Small across all evaluation metrics, even with 50\% fewer parameters in the UNet compared to BK-SDM Small as discussed with~\cref{tab:model_compression}. 
It is important to note that while BK-SDM Small benefits from teacher weights and real images, C-Micro is trained from random initialization without use of any real images.
Finally, B-Base shows significant improvements across all three metrics over BK-SDM Base with the same architecture, and even approaches the performance levels of the teacher model (SDM-v1.4). \cref{fig:comp_others} compares the generated images of our B-Base and C-Micro with those of the teacher model, BK-SDM Base and BK-SDM Tiny.
Notably, C-Micro demonstrates high-quality images despite its compact size.
Compared to other large diffusion models that rely on hundreds of millions of training images, our models achieve comparable performance with far fewer parameters and without using any real images, by distilling knowledge from a well-trained teacher model.
These results underscore the effectiveness of our method, providing an efficient compression solution that maintains high performance.

\usetikzlibrary{shapes.geometric}

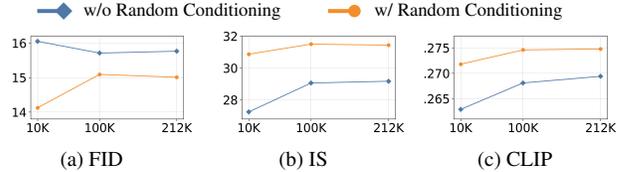
\begin{figure}[t]
    \centering
    \begin{tikzpicture}
        \draw[line width=1.5pt, color=bkcolor] (-2em,0) -- ++(1em,0);
        \node[diamond, fill=bkcolor, scale=0.5] at (-1em,0) {};
        \node[font=\scriptsize, right=0.2em] at (-0.8em,0) {w/o Random Conditioning};
        \draw[line width=1.5pt, color=ourscolor] (9em,0) -- ++(1em,0);
        \node[circle, fill=ourscolor, scale=0.5] at (10em,0) {};
        \node[font=\scriptsize, right=0.2em] at (10.2em,0) {w/ Random Conditioning};
    \end{tikzpicture}
    
    \captionsetup[subfigure]{labelformat=parens }

    \begin{subfigure}[b]{0.155\textwidth}
        \centering
        \includegraphics[width=\textwidth]{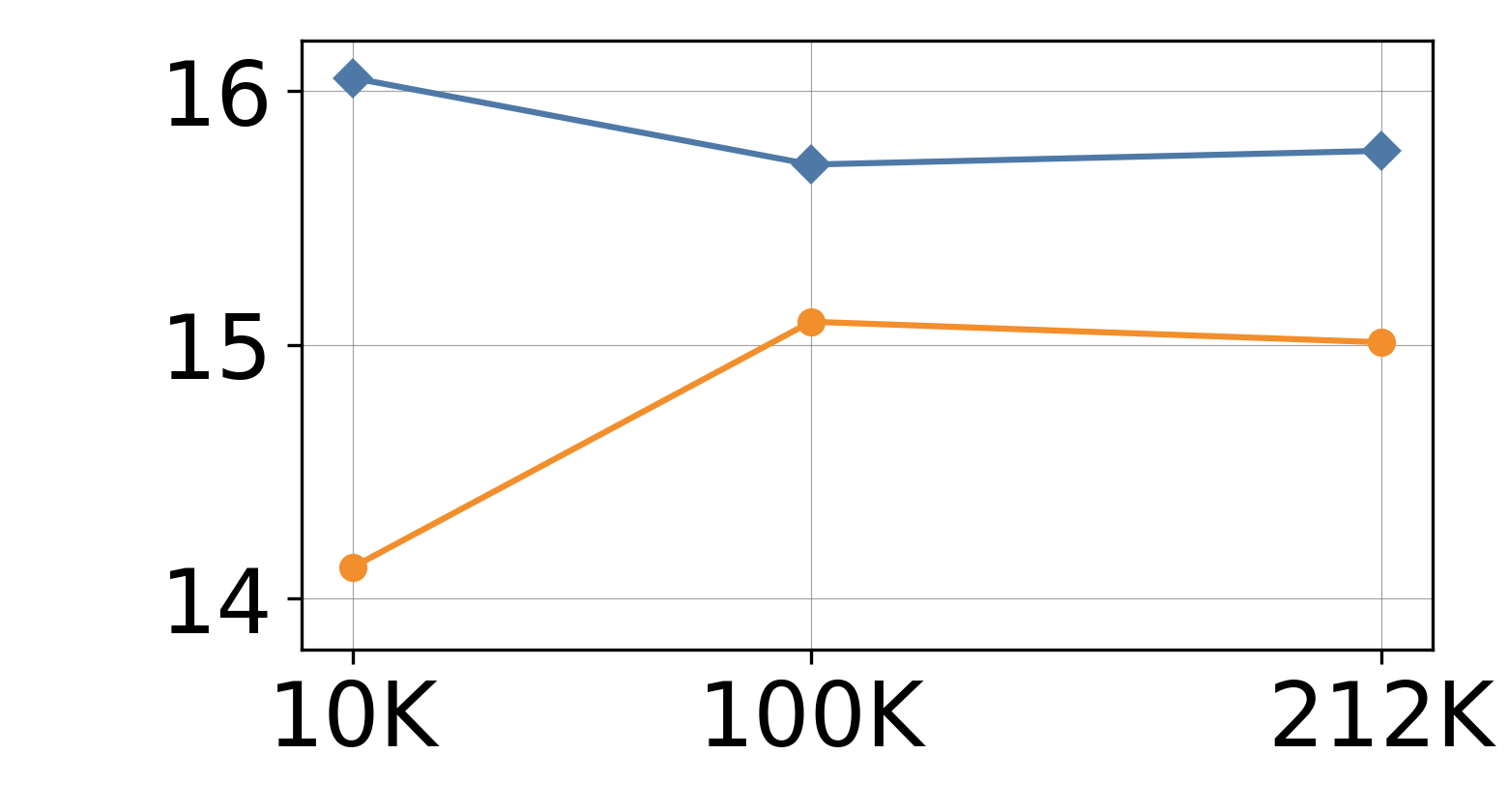}
        \caption{FID}
        \label{fig:sub1}
    \end{subfigure}
    \hfill
    \begin{subfigure}[b]{0.155\textwidth}
        \centering
        \includegraphics[width=\textwidth]{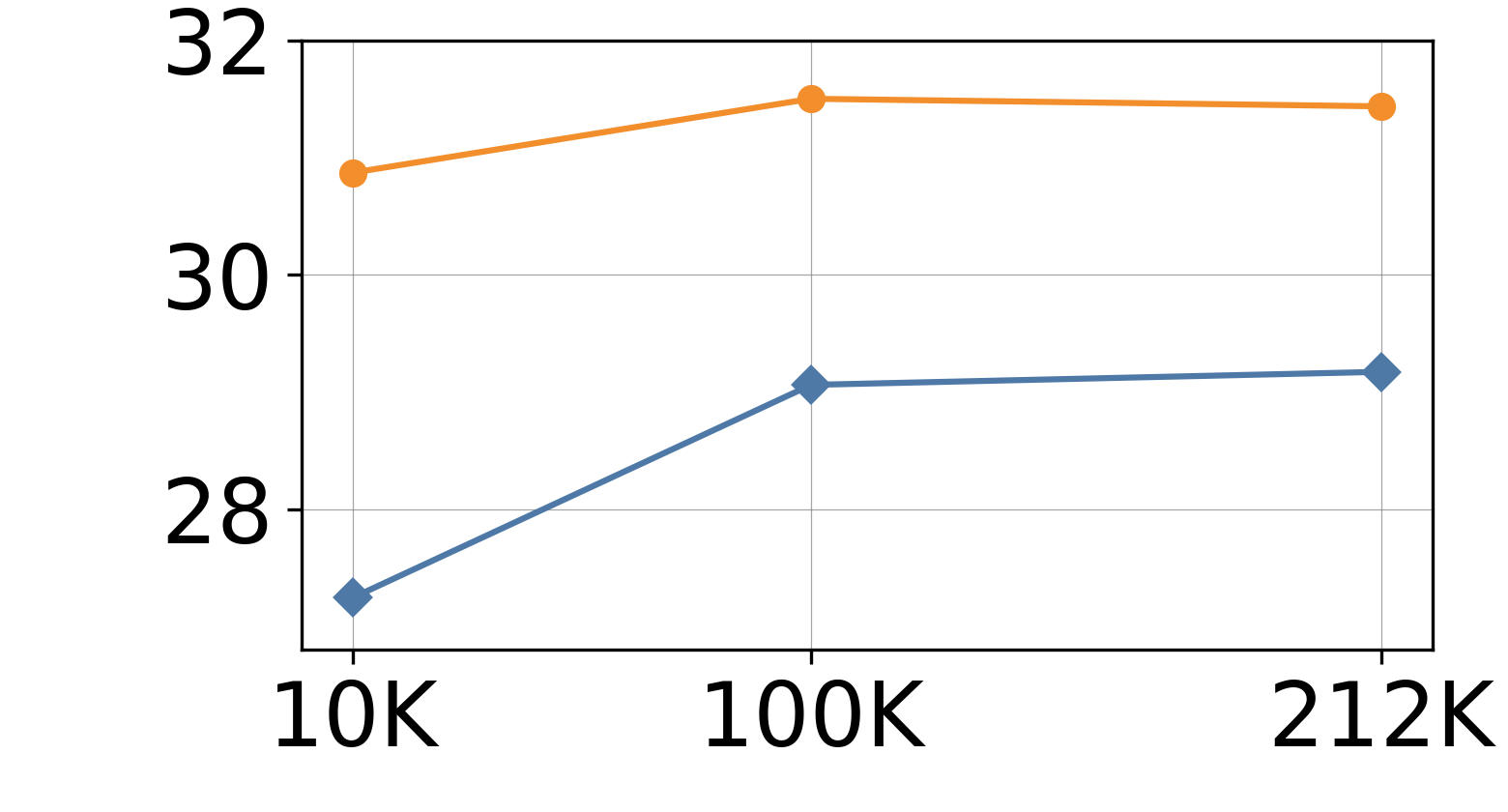}
        \caption{IS}
        \label{fig:sub2}
    \end{subfigure}
    \hfill
    \begin{subfigure}[b]{0.155\textwidth}
        \centering
        \includegraphics[width=\textwidth]{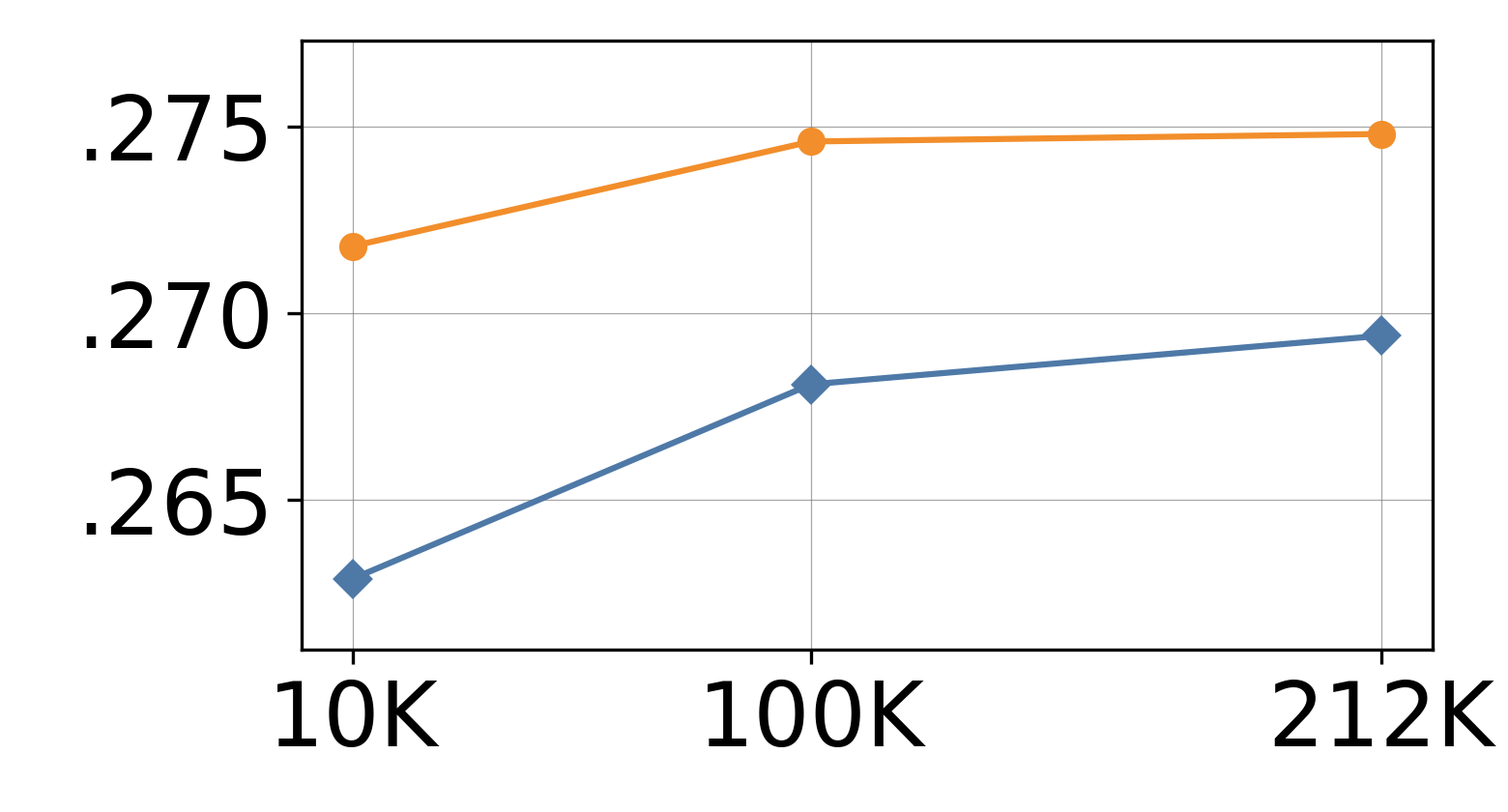}
        \caption{CLIP}
        \label{fig:sub3}
    \end{subfigure}

    \vspace{-0.2cm}
    \caption{\textbf{Impact of Random Conditioning by Cache Size.} We evaluate the models trained with and without random conditioning, using different cache sizes of 10k, 100k, and 212k. 
    We test with B-Base architecture with 125K training iterations.}
    \label{fig:cache_size_ablation}
    \vspace{-0.5em}
\end{figure}

\noindent \textbf{Impact of the Number of Generated Images} \ \ 
In~\cref{fig:cache_size_ablation}, we compare B-Base models with and without random conditioning across different numbers of generated images: 10K, 100K, and 212K.
Across FID, IS, and CLIP scores, models trained with random conditioning consistently outperform those without it. 
Notably, the performance gap widens with fewer generated images (\eg, 10K), showing that random conditioning enables effective distillation even with limited data.
Remarkably, the model trained on 10K images with random conditioning outperforms the one trained on 212K images without it, highlighting the strength of the proposed approach.

\section{Conclusion}
Our work shows that random conditioning enables the student model to learn to generate images of concepts beyond those present in the training image dataset. This capability allows the student to explore a wide text condition space during conditional diffusion model distillation, enhancing performance. This method effectively compresses large diffusion models into smaller, efficient versions.
Additionally, our development of a compact base diffusion model supports use in resource-limited settings and encourages further research advancements.
In this work, the teacher model employed in our experiments is based on Stable Diffusion v1.4. 
We expect that using more advanced versions, such as SDXL, would lead to improved performance due to their enhanced capabilities.
Although our random conditioning method is broadly applicable to distilling conditional diffusion models, our experiments were conducted exclusively on text-to-image models. 
To generalize our findings, future works include extending this approach to diffusion models for other modalities.

\section*{Acknowledgements}
This research was supported by the IITP grants (IITP-2025-RS-2020-II201819, IITP-2025-RS-2024-00436857, RS-2024-00398115), the NRF grants (NRF-2021R1A6A1A03045425) and the KOCCA grant (RS-2024-00345025) funded by the Korea government (MSIT, MOE and MSCT).

{
    \small
    \bibliographystyle{ieeenat_fullname}
    \bibliography{main}
}

\clearpage
\maketitlesupplementary
\appendix

\renewcommand{\thetable}{\Alph{table}}
\renewcommand{\thefigure}{\Alph{figure}}
\setcounter{table}{0}
\setcounter{figure}{0}
\section{Data-Efficient Distillation}
\label{Appendix:Data-Efficient_Distillation}

In our experiments in~\cref{tab:gpt_training}, we highlight our method’s effectiveness under a fully data-free scenario. To address this challenging setup, we develop a systematic workflow for generating a large-scale caption dataset, relying entirely on LLM-generated captions for both image synthesis and random conditioning.

The general workflow consists of two main stages. The first stage involves identifying a comprehensive set of conceptually meaningful \textit{visual} nouns from a structured linguistic database. 
To exclude nouns that are not visual for text-to-image tasks (\eg, abstract concepts such as “justice” or “freedom”), LLM-based filtering step is applied, ensuring only clearly visualizable concepts are selected. 
In the second stage, diverse and extensive textual prompts are systematically generated for each selected noun through LLM inference, resulting in a rich synthetic caption dataset suitable for image synthesis under fully data-free conditions.

Specifically, in our experimental setup, we first extract nouns from WordNet~\cite{miller-1994-wordnet}, restricting the selection to those with a hierarchical depth of 14 to avoid overly specific nouns (\eg, detailed species or subspecies). 
Subsequently, we prompt GPT-3.5-turbo~\cite{gpt35-turbo} to filter only visual nouns, ultimately selecting approximately 17K nouns.
For each selected noun, GPT-3.5-turbo is prompted to generate 128 unique textual prompts, resulting in approximately 2.2M synthetic captions. 
From this pool, we randomly sample 212K prompts, ensuring a balanced distribution across nouns, and utilize a teacher model to generate corresponding image data. Below are the specific GPT prompts used for noun filtering and prompt generation.

\begin{figure}[htbp]
    \captionsetup{labelformat=empty}
    \centering
    \includegraphics[width=1.0\linewidth]{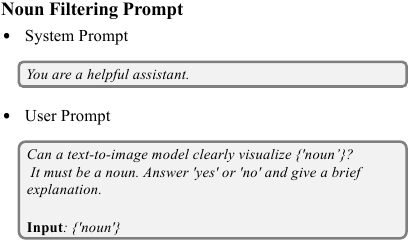}
    \label{fig:prompt2}
\end{figure}

\begin{figure}[htbp]
    \captionsetup{labelformat=empty}
    \centering
    \includegraphics[width=1.0\linewidth]{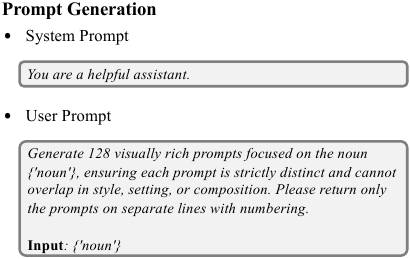}
    \label{fig:prompt3}
\end{figure}

\section{Model Compression}
\label{sup compression}
We evaluate the efficiency of our block- and channel-compressed models in terms of UNet parameter counts, multiply-accumulate operations (MACs), and metric scores, as summarized in~\cref{tab:model_compression}.
The results highlight the efficiency gains of our compressed models with random conditioning over the teacher model while maintaining performances.
\begin{table}[t]
    \centering
    \scalebox{0.85}{
    \setlength{\tabcolsep}{3.4pt}
    \begin{tabular}{lccccc}
        \hline
        \hline
        Models & \#Params & MACs & FID↓ & IS↑ & CLIP↑ \\
                                      
        \hline
        \rowcolor{gray!25}
        Teacher & 860M(+00.0\%)& 339G(+00.0\%)& 13.05&	36.76& 0.2958\\
        
        \hline
        B-Base &580M(-32.6\%) & 224G(-33.9\%) &14.47 & 36.50 & 0.2932\\
        B-Small &483M(-43.9\%)& 218G(-35.7\%) &16.22& 35.99 & 0.2804\\
        B-Tiny &324M(-62.4\%)& 205G(-39.5\%) & 16.71 & 35.46 & 0.2782\\
        \hline
        C-Base & 554M(-35.8\%)& 217G(-36.0\%) &14.45 & 34.92 & 0.2904\\
        C-Small & 426M(-50.5\%) & 166G(-51.0\%) & 14.43&34.58 & 0.2888\\
        C-Tiny& 315M(-63.5\%) & 122G(-64.0\%) & 13.90 & 33.18 & 0.2860\\
        C-Micro & 220M(-74.4\%) & 85G(-74.9\%) & 13.42 & 32.64 & 0.2813\\
        \hline
        \hline
    \end{tabular}
    }
    
    \vspace{-0.2cm}
    \caption{\textbf{Comparison of Model Size and MACs.} We measure UNet parameter count and the MACs for a single step in the UNet. for the teacher model and our models. THOP \cite{Thop} is used to measure the MACs, following the approach of  BK-SDM \cite{kim2023bksdm}. }
    \label{tab:model_compression}
\end{table}

Notably, channel-compressed models show lower MACs than block-compressed models with similar parameter counts; C-Micro requiring only 25\% of the MACs compared to the teacher model.
Despite being trained from scratch, these channel-compressed models demonstrate competitive performance, even outperforming their block-compressed counterparts in several metrics.
This is due to random conditioning, which expands the exploration of the condition space during distillation, offsetting the lack of teacher weight initialization.

\begin{table}[t]
    \centering
    \scalebox{0.85}{
    \begin{tabular}{cccccc}
        \hline
        \hline
        \# & Rand Cond & Additional Text & FID↓ & IS↑ & CLIP↑ \\
        \hline
        1 & \ding{55}  & - & 18.31 & 26.83 & 0.2579 \\
        2 & \ding{51}  & 0 & 16.70 & 26.53 & 0.2613 \\
        3 & \ding{51}  & 1M & 16.13	& 28.89 & 0.2677 \\
        4 & \ding{51}  & 10M & 15.67 & \textbf{28.90} & 0.2674\\
        5 & \ding{51}  & 20M & \textbf{15.22} & 28.89 & \textbf{0.2680}\\
        \hline
        \hline
    \end{tabular}
    }
    \caption{\textbf{Impact of Random Conditioning by Additional Text Size.} We evaluated models trained with additional text sizes of 0, 1M, 10M, and 20M, using the C-Micro architecture for 125K training iterations.
    }
    \label{tab:text_size_abl}
\end{table}
\section{Impact of Additional Text Data Size}
\label{text size}
\cref{tab:text_size_abl} presents the performance results based on the amount of additional text dataset used for random conditioning. Row 1 shows the results of a na\"ive distillation approach without random conditioning, exhibiting lower performance compared to Rows 3, 4, and 5, which incorporate additional text datasets through random conditioning. Notably, Row 2, which applies random conditioning using only the training image-text pairs (212K) without any additional text data (0 additional text), achieves comparable or higher performance compared to Row 1. This indicates that random conditioning effectively utilizes unpaired text-image data during distillation without any performance degradation and can even enhance training.
When comparing Rows 3, 4, and 5, which use additional text datasets of 1M, 10M, and 20M respectively, FID shows a slight improvement as the amount of text data increases, while IS and CLIP scores remain similar. This demonstrates that increasing the amount of text data can enhance training, but even a limited amount, such as 1M, is sufficient to significantly improve model performance. Furthermore, 1M text data requires substantially less memory compared to image datasets and is easier to obtain, highlighting the practical effectiveness of random conditioning in real-world scenarios.

\begin{table*}[t]
    \centering
    \renewcommand{\arraystretch}{1.8}
    \scalebox{0.85}{
    \begin{tabular}{cccccccc}
        \hline
        \hline
        &&
        \multicolumn{3}{c}{Random Initialization} & \multicolumn{3}{c}{Teacher Initialization} \\
        \cline{3-8}
        \# & $p(t)$ & FID↓ & IS↑ & CLIP↑ & FID↓ & IS↑ & CLIP↑ \\
        \hline
        1 & - & 19.69 & 28.75 & 0.2618 & 15.38 & 34.59 & 0.2905\\
        2 & $p(t) = e^{-\lambda \cdot \left(1 - \frac{t}{T}\right)}$ & 16.19& 31.81 & 0.2727 & 14.47 & 36.50 & \textbf{0.2932}\\
        3 & \rule{0pt}{4ex}$p(t) = \begin{cases} 
            e^{-\lambda \cdot \left(1 - \frac{t}{T}\right)}, & \text{if } t > \frac{T}{2} \\[2pt]
            e^{-\lambda \cdot \frac{t}{T}}, & \text{otherwise}
        \end{cases}$ & 15.39 & 31.22 & 0.2730 & 14.53 & \textbf{36.52} & 0.2917\\
        4 & $p(t)=\frac{t}{T}$ & 15.30 & 31.82 & 0.2728 & 14.06 & 36.35 & 0.2898\\
        5 &$p(t) = \frac{1}{1 + e^{-20 \left( \frac{t}{T} - 0.7 \right)}}$ & 15.87 & \textbf{33.34} & \textbf{0.2751} & 15.46 & 35.85 & 0.2916 \\
        6 & $p(t)=0.5$ & 14.68	& 30.70 & 0.2692 & 14.56 & 35.86 & 0.2901\\
        7 & $p(t)=1.0$ & \textbf{14.43} & 28.55 & 0.2586 & \textbf{12.63} & 35.96 & 0.2909\\
        \hline
        \hline
    \end{tabular}
    }
    \caption{\textbf{Effect of $p(t)$ in Random Conditioning.} We evaluate the models trained with varying random conditioning probabilities, $p(t)$. The first row represents the baseline without random conditioning. ``Random Initialization'' and ``Teacher Initialization'' refer to student models trained from scratch and from teacher-initialized weights, respectively.
    All models with random initialization are trained for 125K iterations, while those with teacher-initialized are trained for 75K iterations using the B-Base architecture.}
    \label{tab:rc_rates}
\end{table*}
\begin{figure}[t]
    \centering
    \includegraphics[width=1.0\linewidth]{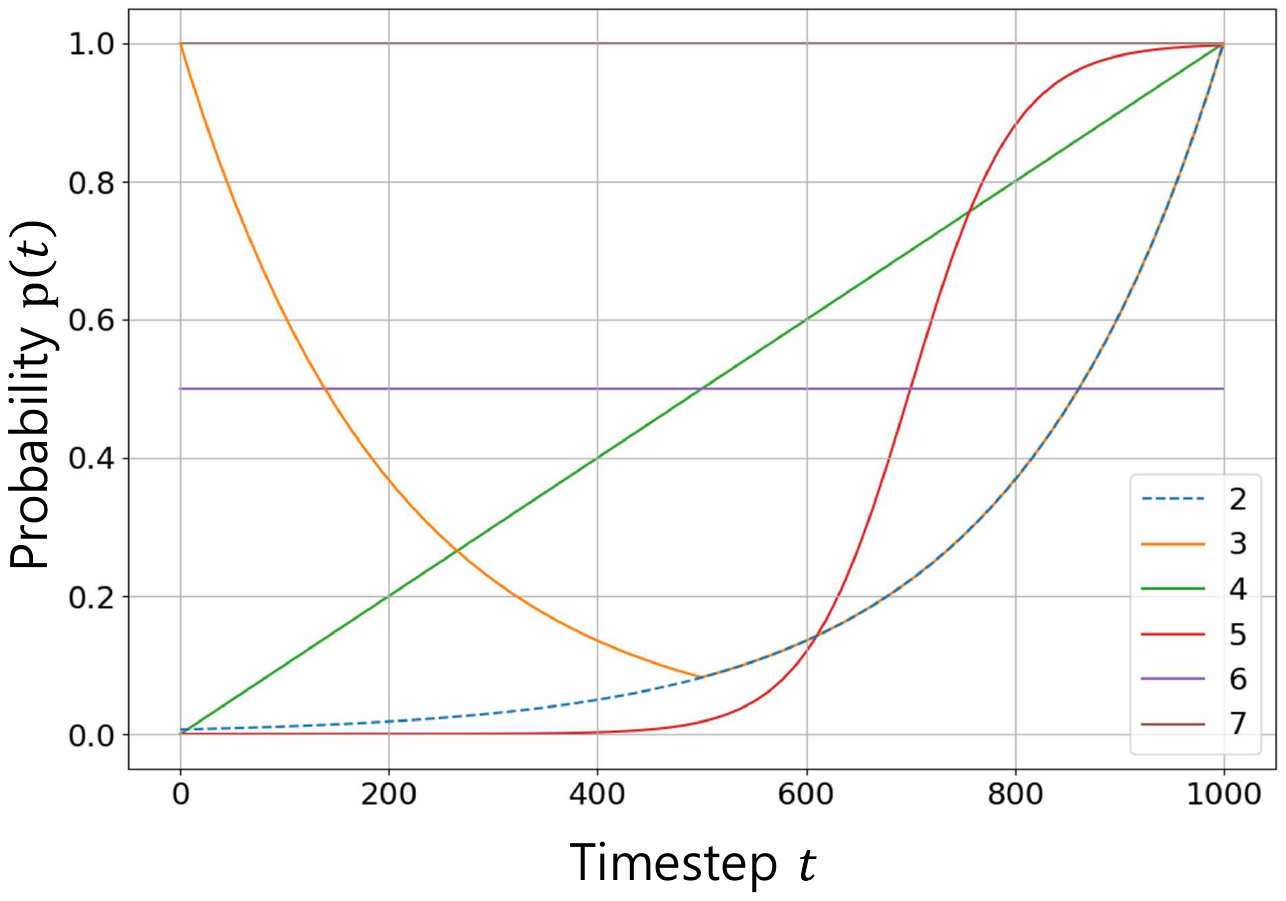}
    \caption{\textbf{Plots of Different $p(t)$ used for Random Conditioning.} Each plot corresponds to a row in \cref{tab:rc_rates}.}
    \label{fig:P_functions}
\end{figure}
\begin{table*}[t]
    \centering
    \scalebox{0.83}{
    \begin{tabular}{ccccccccccccc}
        \hline
        \hline
         & & & & & & \multicolumn{3}{c}{Attribute Binding} & \multicolumn{2}{c}{Object Relationship} & \multirow{2}{*}{Complex↑} & \multirow{2}{*}{Average↑}  \\
         \cmidrule(lr){7-9} \cmidrule(lr){10-11}
        \# & Rand Cond & Real image & FID↓ & IS↑ & CLIP↑ & Color↑& Shape↑ & Texture↑ & Spatial↑ & Non-Spatial↑ & &\\
        \hline
        \rowcolor{gray!25} 
        \multicolumn{3}{c}{(Teacher)} & 13.05 & 36.76 & 0.2958 & 0.3599 & 0.3542 & 0.4004 & 0.1055 & 0.3095 & 0.3095 & 0.3065 \\
        \hline
        1 & \ding{55} & \ding{55} & 18.15 & 33.81 & 0.2864 & 0.3474 & 0.3448 & 0.3786 & 0.0966 & \textbf{0.3092} & 0.3118 & 0.2985\\
        2 & \ding{55} & \ding{51} & 15.76 & 33.79 & 0.2878 & 0.3585 & 0.3397 & 0.3838 & 0.0981 & 0.3080 & 0.3124 & 0.3001 \\
        3 & \ding{51} & \ding{55} & 15.76 & 36.03 & 0.2896 & 0.3593 & 0.3551 & 0.4053 & 0.0889 & 0.3086 & 0.3148 & 0.3053\\
        4 & \ding{51} & \ding{51} & \textbf{15.00} & \textbf{36.14} & \textbf{0.2933} & \textbf{0.3789} & \textbf{0.3576} & \textbf{0.4207} & \textbf{0.1112} & 0.3075 & \textbf{0.3196} & \textbf{0.3159} \\
        \hline
        \hline
    \end{tabular}
    }
    \caption{\textbf{Results on T2I-CompBench~\cite{huang2023t2icompbench} with Additional Metrics.} The row with a gray background shows the performance of the teacher model~\cite{Sdm_v1.4} for reference. “Rand Cond” indicates whether random conditioning is applied, and “Real image” specifies the use of real images during training. All models are based on the B-Base architecture.} 
    \label{tab:sup_compbench}
\end{table*}
\section{Random Conditioning Probability}
\label{sup p test}
\cref{tab:rc_rates} shows the scores for different random conditioning probabilities in two different setups: student models with random and teacher initialization.
The different $p(t)$ functions are plotted in \cref{fig:P_functions}. 
Row~2 corresponds to the exponential function used in the main  experiments. Row~3 modifies this by symmetrically mirroring the function around $t=T/2$. Row~4 represents a linear function that increases from 0 at $t=0$ to 1 at $t=T$. Row~5 uses a sigmoid function, shifted horizontally to ensure $p(t)$ approaches 1 for large time steps. Rows~6 and 7 employ constant functions, with probabilities fixed at 0.5 and 1, respectively, for all $t$.
When the student is randomly initialized, all $p(t)$ functions except for $p(t)=1$ outperform the baseline without random conditioning (Row~1). In particular, the sigmoid function (Row 5) yields the strongest performance. 
When the model is initialized with the teacher weights, the improvements are less pronounced compared to Random Init, but most $p(t)$ functions still lead to better performance than the baseline, with the exponential function (Row 2) showing the best results.

Based on these observations, the final models reported in~\cref{table:main_comp} and~\cref{tab:model_compression} use the sigmoid function (Row 5) for C-compressed models (Random Initialization) and the exponential function (Row 2) for B-compressed models (Teacher Initialization), as these $p(t)$ choices led to strong CLIP and IS scores in each setting. For all other experiments, we use the exponential function (Row 2) as the default $p(t)$.
While this choice may not be optimal, it demonstrates a significant performance improvement over the baseline, validating its effectiveness.
\section{Discussions on Other Image-Free Methods.}
There exist several image-free distillation approaches for building one-step diffusion models~\cite{BOOT, Scoreidentity, SiD-LSG, swiftbrush}, which do not require learning from intermediate noisy samples $\bx_t$ since one-step models directly generate outputs without denoising steps.
These methods receive signals from a pretrained teacher model during training, making them independent of explicit dataset requirements.
However, in our case, the objective is to compress the teacher model into a smaller model while preserving its characteristics, which necessitates training on intermediate noisy samples $\bx_t$.
Similarly, DKDM~\cite{DKDM} applies Dynamic Iterative Distillation for efficient compression, and although it also generates $\bx_t$ through sampling from the teacher model, it specifically targets unconditional models.
As a result, these methods are not directly applicable to our target task—distilling a conditional diffusion model with a large conditioning space using only a limited number of generated images.

\begin{table}[t]
    \centering
    \scalebox{0.85}{
    \begin{tabular}{cccccc}
        \hline
        \hline
        \# & Rand Conditioning & Feature loss &FID↓ & IS↑ & CLIP↑ \\
        \hline
        1 &\ding{55}&\ding{55}& 15.96 & 33.30 & 0.2786 \\
        2 &\ding{55}&\ding{51}& 18.15 & 33.81 & 0.2864 \\
        \hline
        3 &\ding{51}&\ding{55}& \textbf{13.71} & 34.10 & 0.2847 \\
        4 &\ding{51}&\ding{51}& 15.76 & \textbf{36.03} & \textbf{0.2896} \\
        \hline
        \hline
    \end{tabular}
    }
    \vspace{-0.3cm}
    \caption{\textbf{Impact of Feature Loss.} We evaluate the effect of random conditioning with and without feature loss. All models are based on the B-Base architecture.}
    \label{tab:feat_loss}
\end{table}

\section{Inconsistent Trends on FID}
\label{sec: FID inconsistency}
Although random conditioning generally improves performance across FID, IS, and CLIP scores, the trend is not always consistent for FID. As noted in~\cite{huang2023t2icompbench}, FID is known to exhibit substantial fluctuations, making it less reliable for fine-grained comparison. To better validate our findings, we additionally evaluate models on T2I-CompBench~\cite{huang2023t2icompbench}. The results, presented in~\cref{tab:sup_compbench}, show improvements across most reported metrics, aligning well with IS and CLIP scores. These findings support our hypothesis that random conditioning helps the student model better explore the condition space, leading to performance improvements.

\section{Significance of Feature Losses}
In Table \ref{tab:feat_loss}, we report additional experiments comparing the performance of our method with and without feature losses. By comparing Row 1 and Row 3, we observe that even in the absence of feature losses, random conditioning remains effective. Moreover, the comparisons between Row 1 and Row 2 as well as between row3 and Row 4 indicate that incorporating feature losses leads to improvements in both the IS and CLIP scores. These results are consistent with the findings in~\cite{kim2023bksdm}. Note that our feature loss implementation follows those in~\cite{kim2023bksdm}, and we also observed a tendency for lower FID scores when feature losses are omitted.

\begin{table}[t]
  \centering
\scalebox{0.85}{
  \begin{tabular}{cccccc}
    \hline
    \# & Random Conditioning & \#Params &  FID↓ & IS↑ & CLIP↑ \\
    \hline
    \rowcolor{gray!25}     
    \multicolumn{2}{c}{(Teacher)} & 2.56B & 13.04 & 35.83 & 0.3257 \\
    1 &\ding{55} & 0.78B & 23.28 & 27.93 & 0.2855 \\
    2 &\ding{51} & 0.78B & \textbf{21.45} & \textbf{28.53} & \textbf{0.2905} \\ 
    \hline
  \end{tabular}
  }
  \caption{\textbf{Effect of Random Conditioning in SDXL.}
  The row with a gray background corresponds to the Teacher model~(SDXL-Base~\cite{SDXL}), while Rows~1 and~2 represent student models trained using the KOALA-700M architecture.}
  \vspace{-0.1cm}
  \label{tab:sup_sdxl}
\end{table}
\section{SDXL Compression with Koala}
We apply random conditioning while compressing SDXL using the KOALA-700M~\cite{Lee@koala} architecture as the student model, distilling from the SDXL-Base model~\cite{SDXL}. Due to resource constraints, the number of training iterations is lower than what is reported for KOALA, resulting in relatively lower scores. The results are presented in~\cref{tab:sup_sdxl}. 
Following the main experiments, we use 212K training images generated by the teacher model (SDXL-Base) using captions from LAION-Aesthetics V2 (L-Aes) 6.5+. We also incorporate an additional 20M text dataset through random conditioning.

\begin{table}[t]
  \centering
\scalebox{0.85}{
  \begin{tabular}{ccccc}
    \hline
    Method & Random Conditioning & FID↓ & IS↑ & CLIP↑ \\
    \hline
    SLIM & \ding{55} & 27.79 & 16.76 & 0.2063 \\
    SLIM & \ding{51} & 23.97 & 18.83 & 0.2174 \\ 
    \cdashline{1-5}
    BK-SDM & \ding{55} & 15.76 & 33.79 & 0.2878\\
    BK-SDM & \ding{51} & \textbf{15.00} & \textbf{36.14} & \textbf{0.2933}\\
    \hline
  \end{tabular}
  }
  \caption{\textbf{Effect of Random Conditioning in SLIM.}
  We evaluate the impact of random conditioning within SLIM’s loss function and architecture. The “Method” column indicates the model configuration. All models are based on the B-Base architecture.}
  \vspace{-0.1cm}
  \label{tab:sup_slim}
\end{table}
\section{SLIM-Based Distillation}
We also apply random conditioning in a distillation setup following SLIM's~\cite{slim} loss function and architecture. Specifically, we use the authors’ code to compress Stable Diffusion 1.4 on the B-Base architecture, incorporating the Dynamic Wavelet Gating module and using frequency loss. Across all tested configurations, models trained with Random Conditioning consistently achieve higher performance. However, in the SLIM setting, factors such as the smaller dataset size (212K vs. 400M in the original paper), fewer feature losses, and SLIM’s unique model structure make it less competitive in our configuration. The results are shown in~\cref{tab:sup_slim}.
We adopt the same dataset setup as the main experiments.

\section{Details of Animal-Related Data Filtering}
\label{sec: detail filtering}

\noindent \textbf{Training set} \ \ 
To exclude animal images from training, we apply a filtering process to the original 212K LAION~\cite{laion_aesthetics} dataset.
First, each caption is checked for animal-related terms using a curated list, expanded from the 10 MS-COCO animal category names via GPT-4o~\cite{gpt4o} prompting and manual review.
Next, the remaining captions are assessed by GPT-3.5-turbo~\cite{gpt35-turbo}, prompted to determine whether they are in any way related to animals.
Finally, to catch cases where original captions miss animal presence, we generate new captions using BLIP~\cite{blip} and apply the same filtering process.

\noindent \textbf{Evaluation set} \ \ 
For the analysis presented in~\cref{tab:seen_unseen}, we use the MS-COCO validation split, which consists of 41K (40,504) image-text pairs. To compare performance across different dataset compositions, we construct two subsets: one containing 8K (8,265) samples categorized under the “animal” supercategory and the other comprising the remaining 33K (32,239) samples.

\begin{figure}[htbp]
    \captionsetup{labelformat=empty}
    \centering
    \includegraphics[width=1.0\linewidth]{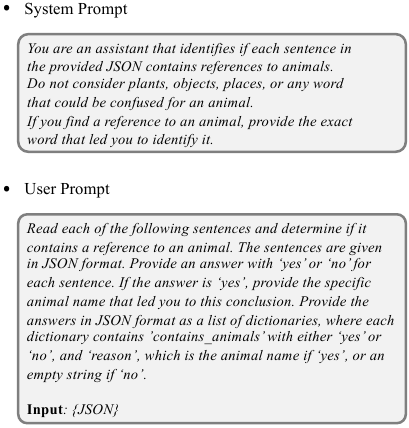}
    \label{fig:prompt1}
\end{figure}

\section{Further Implementation Details}
During inference, including caching $\bx_0$ images for training, evaluation, and generating qualitative results, we consistently employ the DDIM sampler~\cite{ddim} with a total of $T$=25 sampling steps, and the classifier-free guidance~\cite{cfg} is set to its default value of 7.5. In our experiments, we evaluate the models every 25K iterations. Unless specified otherwise, we report the best scores achieved within 125K iterations for experiments initialized with the teacher model and within 400K iterations for those with random initialization; notably, \cref{tab:seen_unseen} presents scores at 500K iterations. For evaluation, we generate images at a resolution of 512×512 and resize them to 256×256, following~\cite{kim2023bksdm}. In~\cref{table:main_comp}, we evaluate \cite{SDXL},~\cite{chen2023pixartalpha}, and~\cite{SD3.5-Medium} using the default settings from the Diffusers library~\cite{von_Platen_Diffusers_State-of-the-art_diffusion}. Images for these models are generated at 1024×1024 resolution and then resized to 256×256 to adhere to our evaluation protocol.

\section{Prompts Used for Qualitative Samples}
\label{detail_prompts}

The first two prompts correspond to (a), while the last three prompts correspond to (b) in~\cref{fig:comp_seen_unseen}. Notably, the prompts in (b) are related to animals.
We use the following prompts for~\cref{fig:comp_seen_unseen} from left to right:

\begin{itemize} 
    \item My favorite landscape I've visited Mount Assinboine Provincial Park Canada.
    \item Storm Over The Black Sea Poster by Ivan Aivazovsky.
    \item \textcolor{red}{Dogs} on Girder Poster.
    \item David Shepherd, \textcolor{red}{Stag}, oil on canvas.
    \item Fotorolgordijn Schildpad Beautiful Green sea \textcolor{red}{turtle} swimming in tropical island reef in hawaii, split over/underwater picture.
\end{itemize} 
We use the following prompts for~\cref{fig:comp_others} from top to bottom:
\begin{itemize} 
    \item Anthropomorphic jackal wearing steampunk armor, beautiful natural rim light, intricate, fantasy, anubis, elegant, hyper realistic, photo realistic, ultra detailed, concept art, octane render, beautirul natural soft rim light, silver details, elegant, ultra detaied, dustin panzino, giger, mucha.
    \item Medium shot side profile portrait photo of a warrior chief, sharp facial features, with tribal panther makeup in blue on red, looking away, werious but clear eyes, 50mm portrait, photography, hard rim lighting photography.
\end{itemize}

\section{Additional Qualitative Results}
\label{sec_sup:qualitative results}
\begin{figure*}[t]
    \centering
    \includegraphics[width=0.85\linewidth]{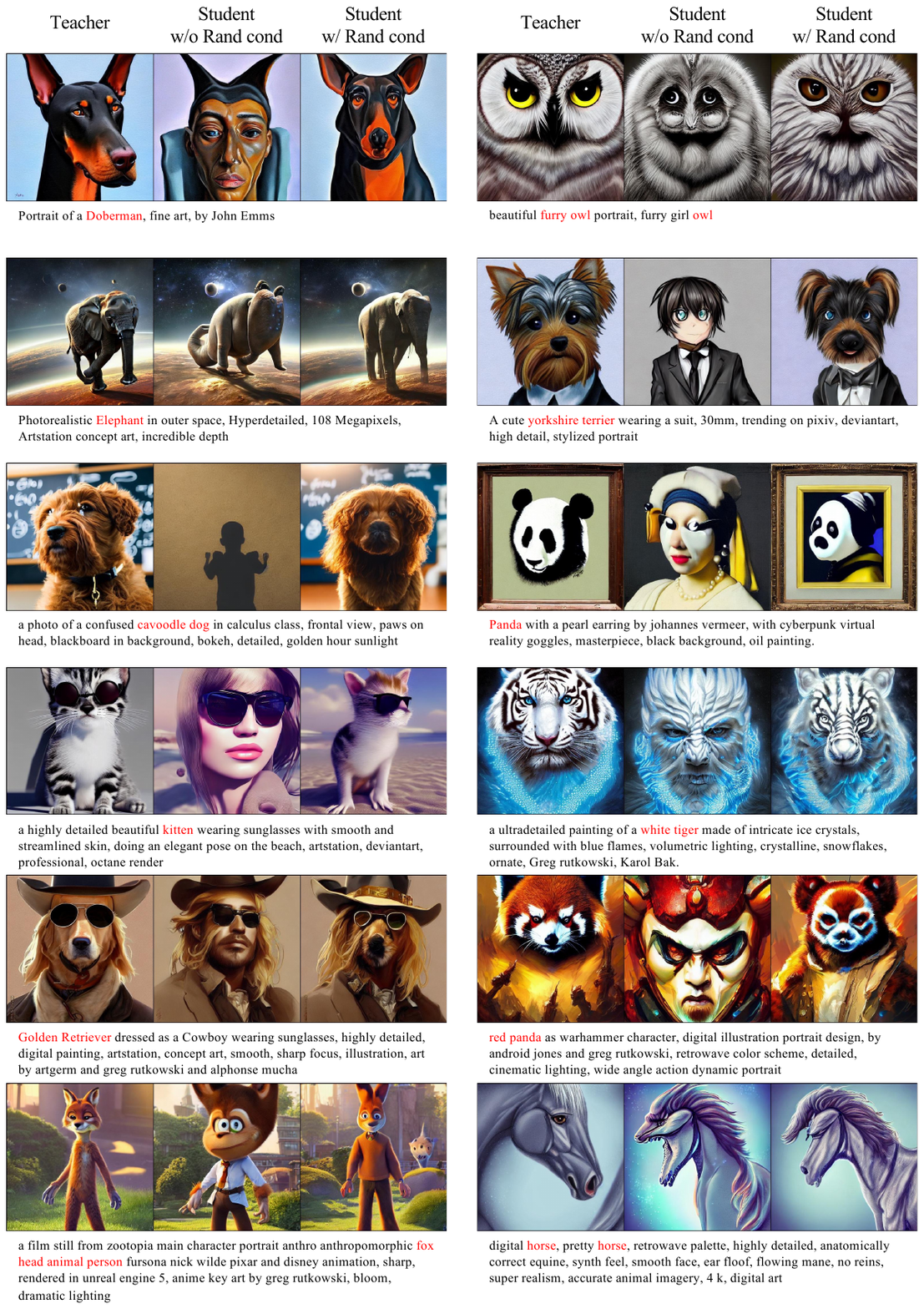}
    \caption{\textbf{Additional Qualitative Results on DiffusionDB \cite{diffusionDB} datasets of Baseline and Our Method Trained Without Animal Image.}
    All samples are generated conditioned on captions related to animals, with animal-related terms highlighted in red for each image's caption}
    \label{fig:appendix_unseen_qual}
\end{figure*}

\begin{figure*}[t]
    \centering
    \includegraphics[width=0.87\linewidth]{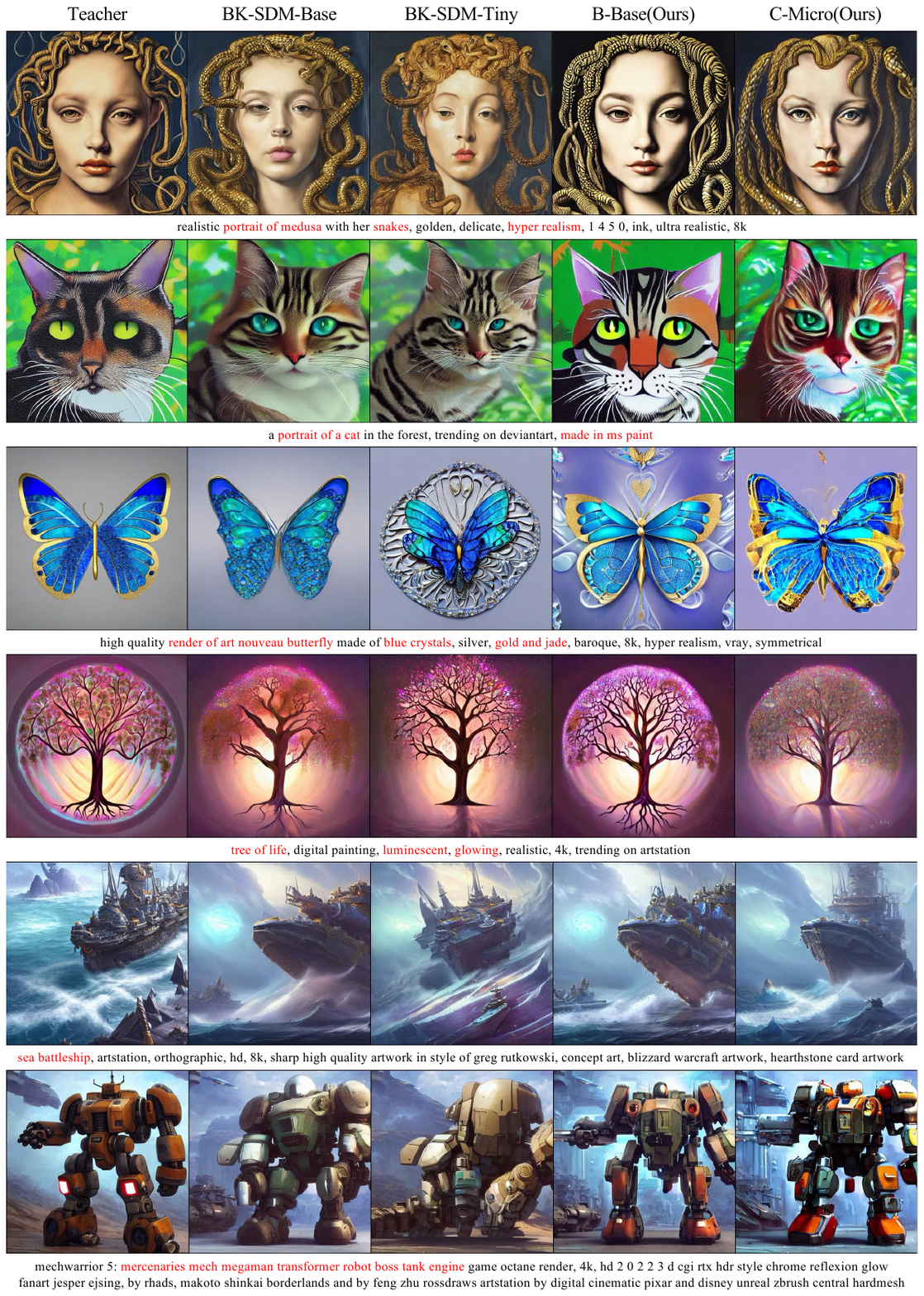}
    \caption{\textbf{Additional Qualitative Results on DiffusionDB \cite{diffusionDB} datasets of Our Models and Baseline Models.}}
    \label{fig:appendix_comp_others}
\end{figure*}

To further illustrate our experiments, we present additional qualitative results generated using the text dataset from DiffusionDB~\cite{diffusionDB}. \cref{fig:appendix_unseen_qual} provides additional results for the animal-related images shown in~\cref{fig:comp_seen_unseen}, while \cref{fig:appendix_comp_others} is corresponding to~\cref{fig:comp_others}.

\end{document}